\documentclass{article}

\usepackage[english]{babel}
\usepackage{float}
\usepackage[letterpaper,top=2cm,bottom=2cm,left=3cm,right=3cm,marginparwidth=1.75cm]{geometry}

\usepackage{url}
\usepackage{mathrsfs, verbatim}
\usepackage{amsthm}
\usepackage{authblk}
\usepackage{amssymb}
\usepackage{float}
\usepackage{graphicx}
\usepackage{algorithm}
\usepackage{algpseudocode}

\usepackage{amsmath}
\usepackage{amsfonts}
\usepackage[colorlinks=true, allcolors=blue]{hyperref}

\newtheorem{lemma}{Lemma}
\newtheorem{theorem}{Theorem}
\newtheorem{proposition}{Proposition}
\newtheorem{definition}{Definition}

\newtheorem{remark}{Remark}

\ifx\assumption\undefined
\newtheorem{assumption}{Assumption}
\fi

\title{DiffFlow: A Unified SDE Framework for Score-Based Diffusion Models and Generative Adversarial Networks}

\date{}
{
\begin{document}

\author[$\ddag$$\dag$]{
Jingwei Zhang
}

\author[$\dag$]{
Han Shi
}

\author[$\ddag\dag$]{
Jincheng Yu
}

\author[$\dag$]{
Enze Xie
}

\author[$\ddag\dag$]{
Zhenguo Li
}

\affil[$\ddag$]{Department of Computer Science and Engineering\\ HKUST}
\affil[$\dag$]{Huawei Noah's Ark Lab}
\affil[ ]{\{jzhangey\}\textit{@}cse.ust.hk}
\affil[ ]{\{shi.han, li.zhenguo\}\textit{@}huawei.com}

\date{}

\maketitle

\begin{abstract}
Generative models can be categorized into two types: explicit generative models that define explicit density forms and allow exact likelihood inference, such as score-based diffusion models (SDMs) and normalizing flows; implicit generative models that directly learn a transformation from the prior to the data distribution, such as generative adversarial nets (GANs). While these two types of models have shown great success, they suffer from respective limitations that hinder them from achieving fast sampling and high sample quality simultaneously.  In this paper, we propose a unified theoretic framework for SDMs and GANs. We shown that: i) the learning dynamics of both SDMs and GANs can be described as a novel SDE named Discriminator Denoising Diffusion Flow (DiffFlow) where the drift can be determined by some weighted combinations of scores of the real data and the generated data; ii) By adjusting the relative weights between different score terms, we can obtain a smooth transition between SDMs and GANs while the marginal distribution of the SDE remains invariant to the change of the weights; iii) we prove the asymptotic optimality and maximal likelihood training scheme of the DiffFlow dynamics; iv) under our unified theoretic framework, we introduce several instantiations of the DiffFLow that provide new algorithms beyond  GANs and SDMs with exact likelihood inference and have potential to achieve flexible trade-off between high sample quality and fast sampling speed.
\end{abstract}

\section{Introduction}
Generative modelling is a fundamental task in machine learning: given finite i.i.d. observations from an unknown target distribution, the goal is to learn a parametrized model that transforms a known prior distribution (e.g., Gaussian noise) to a distribution that is ``close'' to the unknown target distribution. In the past decade,  we have witnessed rapid developments of a plethora of deep generative models (i.e., generative modelling based on deep neural networks): starting from VAEs \cite{kingma2013auto}, GANs \cite{goodfellow2020generative}, Normalizing Flows \cite{rezende2015variational}, and then recent developed score-based diffusion models (SDMs) \cite{sohl2015deep, ho2020denoising}. These deep generative models have shown amazing capability of modelling distributions in high dimensions, which is difficult for traditional ``shallow'' generative models such as Gaussian Mixture Models. 

\noindent Despite a large family of deep generative models has been invented, they are lied in two categories from the perspective of sampling and likelihood inference: explicit generative models are family of generative models that define explicit density forms and thus enable exact likelihood inference \cite{huang2021variational, song2021maximum, kingma2021variational} by the well-known Feynman-Kac formula \cite{karatzas1991brownian}, and typical examples include score-based diffusion models and normalizing flows; implicit generative models such as GANs, on the other hand, directly learn a transformation from the noise prior to the data distribution, making the closed-form density of the learned distribution intractable. In this work, we take GANs in the family of implicit generative models and SDMs in the family of explicit generative models as our objective of study, as they dominate the performance in its respective class of generative models. 

\noindent GANs are trained by playing a minimax game between the generator network and the discriminator network. It is one of the first well-known implicit generative models that dominants the image generation task for many years. The sampling process of GANs is fast since it only requires one-pass to the generator network that transforms the noise vector to the data vector. The downside of GANs, however, is that the training process is unstable due to the nonconvex-nonconcave objective function and the sample quality is inferior compared to the current state-of-the-art score-based diffusion models \cite{dhariwal2021diffusion}. Different from GANs, SDMs achieve high quality image generation without adversarial training. SDM \cite{songscore} is an explicit generative model that defines a forward diffusion process that iteratively deteriorates the data to random noise and the learning objective is to reverse the forward diffusion process by a backward denoising process. The equivalence between denoising and score matching is already known in the literature \cite{hyvarinen2005estimation} and hence it is where the term ``score'' comes from. Nevertheless, the iterative natures make both the training and sampling of SDMs much slower compared to GANs.

\noindent While huge progress has been made in the respective field of GANs and diffusion models, little work has been done on linking and studying the relationship between them. In this work, we aim to answer the following research question:
\begin{center}
\emph{Can we obtain a unified algorithmic and theoretic framework for GANs and SDMs that enables a flexible trade-off between high sample quality and fast sampling speed with exact likelihood inference?}
\end{center}
The goal of this paper is to give an affirmative answer to the above question. The key observation is that the learning dynamics of both SDMs and GANs can be described by a novel SDE named Discriminator Denoising Diffusion Flow (DiffFlow) where its drift term consists of a weighted combination of scores of current marginal distribution $p_t(x)$ and the target distribution. By carefully adjusting the weights of scores in the drift, we can obtain a smooth transition between GANs and SDMs. Here, the we use the word ``smooth'' to imply that the marginal distribution $p_t(x)$ at any time $t$ remains unchanged by the weights adjustments from GANs to SDMs. We name this property the ``Marginal Preserving Property'' which we would give a rigorous definition in later sections. We also provide an asymptotic and nonasymptotic convergence analysis of the dynamics of the proposed SDE under some isoperimetry property of the smoothed target distribution and design a training algorithm with maximal likelihood guarantees. Under our unified algorithmic framework, we can provide several instantiations of DiffFlow that approximate the SDE dynamics by neural networks and enable a flexible trade-off between high sample quality and fast sampling speed.


\section{Background}
\subsection{Score-Based Diffusion Models}
SDMs are type of generative models that are trained by denoising samples corrupted by various levels of noise. Then the generation process is defined by sampling vectors from a pure noise and progressively denoising noise vectors to images. \cite{songscore} formally describes the above processes by a forward diffusion SDE and a backward denoising SDE. 

\noindent Specifically, denote the data distribution by $q(x)$. By sampling a particle $X_0\sim q(x)$, the forward diffusion process $\{X_t\}_{t\in [0, T]}$ is defined by the following SDE:
\begin{eqnarray}
dX_t = f(X_t, t)dt + g(t)dW_t
\end{eqnarray}
where $T>0$ is a fixed constant; $f(\cdot, \cdot): \mathbb{R}^k\times [0, T]\to  \mathbb{R}^k$ is the drift coefficient; 
$g(\cdot):  [0, T]\to  \mathbb{R}_{\geq 0}$ is a predefined diffusion noise scheduler; $\{W_t\}_{t\in [0, T]}$ is the standard Brownian motion in $\mathbb{R}^k$. If we denote the probability density of $X_t$ by $p_t(x)$, then we hope that the distribution of $p_T(x) $ is close to some tractable gaussian distribution $\pi(x)$. If we set $f(X_t, t) \equiv 0$ and $g(t) = \sqrt{2t}$ as in \cite{karras2022elucidating}, then we have $$p_t(x) = q(x) \circledast \mathcal{N}(0, t^2I):= q(x; t)~$$
where $\circledast$ denotes the convolution operation and $q(x, T)\approx \pi(x) = \mathcal{N}(0, T^2I)$.

\noindent  \cite{songscore} define the backward process by sampling an initial particle from $X_0\sim \pi(x)  \approx p_T(x)$. Then the backward denoising process $\{X_t\}_{t\in [0, T]}$ is defined by the following SDE:
\begin{eqnarray}
dX_t = \left[g^2(T-t)\nabla\log p_{T-t}(X_t) - f(X_t, T-t)\right]dt + g(T-t)dW_t~.
\end{eqnarray}
It worths mentioning that the above denoising process is a trivial time-reversal to the original backward process defined in \cite{songscore}, which also appears in previous work \cite{huang2021variational} for notational simplicity. 

\noindent In the denoising process \footnote{We omit the word ``backward'' since this process is the time-reversal of the original backward process.}, the critical term to be estimated is $\nabla\log p_{T-t}(x)$, which is the score function of the forward process at time $T-t$. $p_{T-t}(x)$ is often some noise-corrupted version of the target distribution $q(x)$.  For example, as discussed before, if we set $f(X_t, t) \equiv 0$ and $g(t) = \sqrt{2t}$, then $\nabla\log p_{T-t}(x)$ becomes $\nabla\log q(x, T-t)$, which can be estimated through denoising score matching \cite{songscore, karras2022elucidating} by a time-indexed network $s_{\theta}(x, t)$ \cite{ho2020denoising}.

\subsection{Generative Adversarial Networks}
In $2014$, \cite{goodfellow2020generative} introduce the seminal work of generative adversarial nets. The GAN's training dynamics can be formulated by a minimax game between a discriminator network $d_{\theta_D}(\cdot): \mathcal{X} \to [0, 1]$ and a generator network $G_{\theta_G}(\cdot): \mathcal{Z} \to\mathcal{X}$.  From an intuitive understanding, the discriminator network is trained to classify images as fake or real while the generator network is trained to produce images from noise that ``fool'' the discriminator. The above alternate procedure for training generator and discriminator can be formulated as a minimax game:
\begin{eqnarray}
\min_{\theta_G}\max_{\theta_D} \mathbb{E}_{x\sim q(x)}[\log d_{\theta_D}(x)] +  \mathbb{E}_{z\sim \pi(z)}[\log (1-d_{\theta_D}(G_{\theta_G}(z)) )] ~
\end{eqnarray}
where $z\sim \pi(z)$ is sampled from gaussian noise. The training dynamics of GANs is unstable, due to the highly nonconvexity of the generator and the discriminator that hinders the existence and uniqueness of the equilibrium in the minimax objective. 

\noindent Despite lots of progress has been made in the respective field of diffusion models and GANs in recent years, little is known about the connection between them. In the next section, we will provide a general framework that unifies GANs and diffusion models and show that the learning dynamics of GANs and diffusion models can be recovered as a special case of our general framework. 

\section{General Framework}
In order to obtain a unified SDE framework for GANs and SDMs, let us forget the terminologies ``forward process'' and ``backward process'' in the literature of diffusion models. We aim to construct a learnable (perhaps stochastic) process $\{X_t\}_{t\in [0, T]}$ indexed by a continuous time variable $0\leq t \leq T$ such that $X_0\sim \pi(x)$ for which $\pi(x)$ is a known Gaussian distribution and $X_T \sim p_T(x)$ that is ``close'' to the target distribution $q(x)$. The closeness can be measured by some divergence or metric which we would give details later.

\noindent Specifically, given $X_0\sim \pi(x)$ from noise distribution, we consider the following evolution equation of $X_t\sim p_t(x)$ for $t\in [0, T]$ with some $T>0$:
\begin{eqnarray} \label{DiffFlow}
dX_t =\left[f(X_t, t) + \beta(t, X_t)\nabla\log \frac{q(u(t)X_t; \sigma(t))}{p_t(X_t)}+ \frac{g^2(t)}{2}\nabla\log p_t(X_t)\right]dt  + \sqrt{g^2(t)-\lambda^2(t)}dW_t~
\end{eqnarray}
where $f(\cdot, \cdot): \mathbb{R}^k\times  [0, T]\to \mathbb{R}^{k}$; $ \beta(\cdot, \cdot): \mathbb{R}^k\times  [0, T]\to \mathbb{R}_{\geq 0}$ and $u(\cdot), \sigma(\cdot), g(\cdot), \lambda(\cdot): [0, T]\to \mathbb{R}_{\geq 0}$ are all predefined scaling functions. 

\noindent For the ease of presentation, we name the above SDE the \textbf{``Discriminator Denoising Diffusion Flow'' (DiffFlow)}.  It seems hard to understand the physical meaning behind DiffFlow at a glance, we will explain it in detail how the name DiffFlow comes from and how it unifies GANs and SDMs in later subsections: with careful tuning on the scaling functions, we can recover dynamics of GANs and SDMs respectively without changing the marginal distributions $p_t(x)$. Furthermore, we will see that DiffFlow unifies a wider ``continuous'' spectrum of generative models where GANs and diffusion models are corner cases of DiffFlow with specialized scaling functions.

\subsection{Single Noise-Level Langevin Dynamics}
Let us start with the simplest case: the single-noise level Langevin dynamics. In DiffFlow, if we set $u(t)\equiv 1$, $f(X_t, t) \equiv 0$, $\lambda(t)\equiv 0$, $\beta(t, X_t) \equiv \beta(t)$, $g(t)\equiv \sqrt{2\beta(t)}$ and $\sigma(t)\equiv \sigma_0$ for some fixed $\sigma_0\geq 0$, then we obtain
\begin{eqnarray} \label{DiffFlow}
dX_t =\beta(t)\nabla\log q(X_t; \sigma_0)dt  + \sqrt{2\beta(t)}dW_t~
\end{eqnarray}
which is exactly the classical Langevin algorithm. However, as pointed out by \cite{song2019generative, song2020improved}, the single noise level Langevin algorithm suffers from slow mixing time due to separate regions of data manifold. Hence, it cannot model complicated high dimensional data distribution: it even fails to learn and generate reasonable features on MNIST datasets \cite{song2019generative}. Hence, it is reasonable to perturb data with multiple noise levels and perform multiple noise-level Langevin dynamics, such as the Variance Explosion SDE with Corrector-Only Sampling (i.e., NCSN) \cite{songscore}.

\subsection{Score-based Diffusion Models}

\subsubsection{Variance Explosion SDE}
Without loss of generality, we consider the Variance Explosion (VE) SDE adopted in \cite{karras2022elucidating}. DiffFlow can be easily adapted to other VE SDEs by simply changing the noise schedule $\sigma(t)$ and scaling $\beta(t, X_t)$. We can set $u(t)\equiv 1$, $f(X_t, t) \equiv 0$, $\beta(t, X_t) \equiv 2(T-t)$, $\lambda(t)\equiv \sqrt{\beta(t)} = \sqrt{T-t}$, $g(t)\equiv \sqrt{2\beta(t)} = 2\sqrt{T-t}$ and $\sigma(t)\equiv T-t$, then we obtain
\begin{eqnarray} \label{DiffFlow}
dX_t =2(T-t)\nabla\log q(X_t; T-t)dt  +\sqrt{2(T-t)} dW_t~
\end{eqnarray}
which is the VE SDE in \cite{karras2022elucidating}. For a general VE SDE in \cite{songscore}, the denoising process is 
\begin{eqnarray}
 dX_t = \frac{d[\sigma^2(T-t)]}{dt}\nabla\log q\left(X_t;\sqrt{\sigma^2(T-t)-\sigma^2(0)}\right) + \sqrt{\frac{d[\sigma^2(T-t)]}{dt}}dW_t
\end{eqnarray}
which can be obtained by setting $u(t)\equiv 1$, $f(X_t, t) \equiv 0$, $\beta(t, X_t) \equiv \frac{d[\sigma^2(T-t)]}{dt}$, $\lambda(t)\equiv  \sqrt{\beta(t, X_t)} = \sqrt{\frac{d[\sigma^2(T-t)]}{dt}}  $, $g(t)\equiv \sqrt{2\beta(t, X_t)} = \sqrt{2\frac{d[\sigma^2(T-t)]}{dt}}   $ and $\sigma(t)\equiv \sqrt{\sigma^2(T-t)-\sigma^2(0)}$ in DiffFlow.

\subsubsection{Variance Preserving SDE}
\noindent Similar to previous analysis, for Variance Preserving (VP) SDE in \cite{songscore}, the denoising process is 
\begin{eqnarray}
dX_t = \left[\beta(T-t)\nabla\log p_{T-t}(X_t) +\frac{1}{2} \beta(T-t)X_t\right]dt + \sqrt{\beta(T-t)}dW_t~.
\end{eqnarray}
Now we need to represent the score of the forward process $\nabla\log p_{t}(X_t)$ in VP SDE by the score of the noise-corrupted target distribution. 
We have the following proposition, which can be obtained by simple application of stochastic calculus. 
\begin{proposition}
Assume $X_0\sim q(x)$ and the forward process $\{X_t\}_{t\in [0, T]}$ in VP SDE is given by
\begin{eqnarray}
&& dX_t = -\frac{1}{2}\beta(t)X_tdt + \sqrt{\beta(t)}dW_t~.
\end{eqnarray}
Then the score of $p_t(x)$ (note: $X_t\sim p_t(x)$) can be represented as
\begin{eqnarray}
&&  \nabla \log p_t(x) =  \nabla\log q\left( \exp\left(\frac{1}{2}\int_0^t\beta(s)ds\right) x; 1- \exp\left(-\int_0^t\beta(s)ds\right)\right)~.
\end{eqnarray}
\begin{proof}
By ito lemma, we have
\begin{eqnarray}
&& d\left[\exp\left(\frac{1}{2}\int_0^t\beta(s)ds\right)X_t\right] = \exp\left(\frac{1}{2}\int_0^t\beta(s)ds\right)dX_t +   \frac{1}{2}\beta(t)\exp\left(\frac{1}{2}\int_0^t\beta(s)ds\right)X_tdt~.
\end{eqnarray}
Combining with the forward VP SDE, we have
\begin{eqnarray}
&& d\left[\exp\left(\frac{1}{2}\int_0^t\beta(s)ds\right)X_t\right] = \exp\left(\frac{1}{2}\int_0^t\beta(s)ds\right) \sqrt{\beta(t)}dW_t~.
\end{eqnarray}
Hence,
\begin{eqnarray}
&& X_t = \exp\left(-\frac{1}{2}\int_0^t\beta(s)ds\right)\left[ X_0 + \int_{0}^{t}\exp\left(\frac{1}{2}\int_0^u\beta(s)ds\right) \sqrt{\beta(u)}dW_u \right]~.
\end{eqnarray}
By the ito isometry and martingale property of brownian motions, we have
\begin{eqnarray}
\mathbb{E}\left[\int_{0}^{t}\exp\left(\frac{1}{2}\int_0^u\beta(s)ds\right) \sqrt{\beta(u)}dW_u\right] = 0
\end{eqnarray}
and
\begin{eqnarray}
&& \mathbb{E}\left[\int_{0}^{t}\exp\left(\frac{1}{2}\int_0^u\beta(s)ds\right) \sqrt{\beta(u)}dW_u\right]^2 \nonumber\\
&& = \int_{0}^{t}\left[\exp\left(\frac{1}{2}\int_0^u\beta(s)ds\right) \sqrt{\beta(u)}\right]^2du \nonumber\\
&& = \int_{0}^{t}\exp\left(\int_0^u\beta(s)ds\right) \beta(u)du \nonumber\\
&& = \int_{0}^{t}\exp\left(\int_0^u\beta(s)ds\right) d \left[\int_0^u\beta(s)du \right]  \nonumber\\
&& =  \exp\left(\int_0^t\beta(s)ds\right) -1~. 
\end{eqnarray}
Hence
\begin{equation}
\exp\left(-\frac{1}{2}\int_0^t\beta(s)ds\right) \int_{0}^{t}\exp\left(\frac{1}{2}\int_0^u\beta(s)ds\right) \sqrt{\beta(u)}dW_u \sim \mathcal{N}\left(0,  I- \exp\left(-\int_0^t\beta(s)ds\right)I  \right)~.
\end{equation}
Then we have 
\begin{equation}
p_t(x)  = \exp\left(\frac{1}{2}\int_0^t\beta(s)ds\right)q\left(\exp\left(\frac{1}{2}\int_0^t\beta(s)ds\right) x\right) \circledast  \mathcal{N}\left(0,  I- \exp\left(-\int_0^t\beta(s)ds\right)I  \right)
\end{equation}
where the proof ends by taking logarithm and applying the divergence operator $\nabla$ on both sides. 
\end{proof}
\end{proposition}
\noindent Now, we can obtain the denoising process of the VP SDE as follows, 
\begin{eqnarray}
dX_t = \left[\beta(T-t) \nabla\log q\left( \exp\left(\frac{1}{2}\int_0^{T-t}\beta(s)ds\right) X_t; 1- \exp\left(-\int_0^{T-t}\beta(s)ds\right)\right) +\frac{1}{2} \beta(T-t)X_t\right]dt + \sqrt{\beta(T-t)}dW_t~ \nonumber
\end{eqnarray}
which can be obtained by setting $u(t)\equiv  \exp\left(\frac{1}{2}\int_0^{T-t}\beta(s)ds\right)$, $f(X_t, t) \equiv \frac{1}{2} \beta(T-t)X_t$,  $\beta(t, X_t) \equiv \beta(T-t)$, $\lambda(t)\equiv \sqrt{\beta(T-t)}$,  $g(t)\equiv   \sqrt{2\beta(T-t)}   $ and $\sigma(t)\equiv 1- \exp\left(-\int_0^{T-t}\beta(s)ds\right)$  in DiffFlow. Similar procedure can show that sub-VP SDE proposed in \cite{songscore} is also lied in the DiffFlow framework with specialized scaling functions, we leave it as a simple exercise for readers.

\subsubsection{Diffusion ODE Flow}
Similar to previous analysis, for the diffusion ODE corresponds to VE SDE in \cite{songscore}, the denoising process is 
\begin{eqnarray}
 dX_t = \frac{1}{2}\frac{d[\sigma^2(T-t)]}{dt}\nabla\log q\left(X_t;\sqrt{\sigma^2(T-t)-\sigma^2(0)}\right)
\end{eqnarray}
which can be obtained by setting $u(t)\equiv 1$, $f(X_t, t) \equiv 0$, $\beta(t, X_t) \equiv \frac{1}{2}\frac{d[\sigma^2(T-t)]}{dt}$, $\lambda(t)\equiv  \sqrt{2\beta(t, X_t)} = \sqrt{2\frac{d[\sigma^2(T-t)]}{dt}}  $, $g(t)\equiv \sqrt{2\beta(t, X_t)} = \sqrt{2\frac{d[\sigma^2(T-t)]}{dt}}   $ and $\sigma(t)\equiv \sqrt{\sigma^2(T-t)-\sigma^2(0)}$ in DiffFlow. The ODEs correspond to VP SDEs and sub-VP SDEs can be obtained from DiffFlow by specilizing scaling functions with nearly the same procedure, hence we omit the derivations.

\subsection{Generative Adversarial Networks}
\subsubsection{The Vanilla GAN}
To start with the simplest case, let us show how DiffFlow recovers the training dynamics of the vanilla GAN \cite{goodfellow2020generative}. First, let us set $u(t)\equiv 1$, $f(X_t, t) \equiv 0$,  $\lambda(t)\equiv 0$,  $g(t)\equiv  0  $ and $\sigma(t)\equiv  \sigma_0\geq 0$. Notice that $\sigma_0$ is usually set to be a small positive constant to ensure the smoothness of the gradient for the generator. Then, the DiffFlow reduces to the following DiffODE:
\begin{eqnarray} 
dX_t =\left[ \beta(t, X_t)\nabla\log \frac{q(X_t;\sigma_0)}{p_t(X_t)}\right]dt ~.
\end{eqnarray}
Next, we show that by an extreme coarse approximating to the dynamics of this ODE by a generator network with specialized $\beta(t, X_t)$ , one recovers the dynamics of the Vanilla GAN.  

\noindent Note that the critical term to be estiamted in DiffODE is $\nabla\log \frac{q(X_t;\sigma_0)}{p_t(X_t)}$, which is the gradient field of the classifier between the real data and generated data at time $t$. This term can be estimated by taking gradients to the logistic classifier defined as follows: 
\begin{eqnarray}
&& D_t(x) := \log \frac{q(x;\sigma_0)}{p_t(x)} = \arg\min_{D}\left[\mathbb{E}_{x\sim q(x;\sigma_0)}\log\left(1+e^{-D(x)}\right) + \mathbb{E}_{x\sim p_t(x)}\log\left(1+e^{D(x)}\right) \right]
\end{eqnarray} 
Then we can update the samples by
\begin{eqnarray}\label{GAN}
&& X_{t+1} = X_t + \eta_t\beta(t, X_t) \nabla D_t(X_t) 
\end{eqnarray} 
where $\eta_t>0$ is the discretization stepsize. Hence, DiffODE naturally yield the following algorithm and we can show this algorithm is equivalent to GAN by setting the discriminator loss to logistic loss: $d_{\theta_D}(x) = \frac{1}{1+e^{-D_{\theta_D}(x)}}$.

\begin{algorithm} [H]
        \caption{DiffFlow-GANs}
        \label{alg4}
        \textbf{INPUT: target data from $q(x;\sigma_0)$: $x_1^*,\ldots,x_n^*\in\mathbb{R}^k$; noisy samples $x_0^0, \ldots, x_n^0\in\mathbb{R}^k$ generated from noisy distribution $\pi(x)$; meta-parameter: $T$; neural network classifier: $D_{\theta_D}(x)$; generator: $G_{\theta_G}(x)$.} \vskip 3mm
        \begin{algorithmic} 
        \For  {$t=1,\ldots, T$} \vskip 1mm
        Let $ \theta_D^{t-1} = \arg\min_{\theta_D} \left[ \frac{1}{n}\sum_{i=1}^{n}\log\left(1+e^{-D_{\theta_D}(x_i^*)}\right) + \frac{1}{n}\sum_{i=1}^{n}\log\left(1+e^{D_{\theta_D}(x_i^{t-1})}\right)   \right]$ \vskip 1mm
       Sample $z_1, \ldots, z_n$ from noisy prior $\pi(x)$.  \vskip 1mm
       Update the generator by descending the following loss: $$\frac{1}{2n}\sum_{i=1}^n \|G_{\theta_G}(z_i) - (G_{\theta_G^{t-1}}(z_i)+\eta_t\beta(z_i, t)\nabla D_{\theta_D^{t-1}}(G_{\theta_G^{t-1}}(z_i)))\|_2^2~.$$  \vskip 1mm
       Update the sampled particles: $x_i^{t} = x_i^{t-1} + \eta_t\beta(x_i^{t-1}, t )\nabla_{\theta_D} D_{\theta_D^{t-1}}(x_i^{t-1})$ for  $i=1,\ldots, n$
         \EndFor\\
         \Return $G_{\theta_G^T}(x)$ and particles $\{x_i^{T}\}_{i=1}^n~.$
        \end{algorithmic}
\end{algorithm}

\noindent For the vanilla GANs \cite{goodfellow2020generative}, the update of discriminator is the same as DiffODE-GANs. Since from $d_{\theta_D}(x) = \frac{1}{1+e^{-D_{\theta_D}(x)}}$,  we have
\begin{eqnarray}
&&\mathbb{E}_{x\sim q(x;\sigma_0)}\log\left(1+e^{-D_{\theta_D}(x)}\right) + \mathbb{E}_{x\sim p_t(x)}\log\left(1+e^{D_{\theta_D}(x)}\right)  \nonumber\\
&& = \mathbb{E}_{x\sim q(x;\sigma_0)}\log\left(1+e^{-\log\frac{d_{\theta_D}(x)}{1-d_{\theta_D}(x)}}\right) + \mathbb{E}_{x\sim p_t(x)}\log\left(1+e^{\log\frac{d_{\theta_D}(x)}{1-d_{\theta_D}(x)}}\right) \nonumber\\
&& = -\mathbb{E}_{x\sim q(x;\sigma_0)}\log\left(d_{\theta_D}(x)\right) - \mathbb{E}_{x\sim p_t(x)}\log\left(1-d_{\theta_D}(x)\right) ~.
\end{eqnarray}
 It remains to show that the update of the generator is also equivalent. The gradient of the generator is 
\begin{eqnarray}
&& \nabla_{\theta_G}\log(1-d_{\theta_D}(G_{\theta_G}(z))) \nonumber\\
&& = -  \nabla_{\theta_G}\log(1+e^{D_{\theta_D}(G_{\theta_G}(z))})\nonumber\\
&& = - \frac{1}{1+e^{-D_{\theta_D}(G_{\theta_G}(z))}}\nabla_{\theta_G}D_{\theta_D}(G_{\theta_G}(z))\nonumber\\
&& = - d_{\theta_D}(G_{\theta_G}(z)) \nabla_{\theta_G}D_{\theta_D}(G_{\theta_G}(z))\nonumber\\
&& = - d_{\theta_D}(G_{\theta_G}(z)) \nabla D_{\theta_D}(G_{\theta_G}(z))\cdot \nabla_{\theta_G}G_{\theta_G}(z)~.
\end{eqnarray}
Hence, at the time step $t-1$, we obtain the discriminator with parameter $\theta_D^{t-1}$ and update generator by the following equation
\begin{eqnarray}
&& \theta_G^{t} = \theta_G^{t-1}  + \lambda_t \left[\frac{1}{n}\sum_{i=1}^{n}d_{\theta_D^{t-1}}(G_{\theta_G^{t-1}}(z_i)) \nabla D_{\theta_D^{t-1}}(G_{\theta_G^t}(z_i))\cdot\nabla_{\theta_G}G_{\theta_G^{t-1}}(z)  \right]
\end{eqnarray}
where $z_i\sim\mathcal{N}(0, I)$ and $\lambda_t$ is the learning rate for mini-batch SGD at time $t$.  

\noindent If we instead run a gradient descent step on the MSE loss of the generator in the DiffFlow-GAN, we obtain
\begin{eqnarray}
&& \theta_G^{t} = \theta_G^{t-1}  + \lambda_t\left[\frac{1}{n}\sum_{i=1}^{n}\eta_t \beta(z_i, t)\nabla D_{\theta_D^t}(G_{\theta_G^t}(z_i))\cdot\nabla_{\theta_G}G_{\theta_G^t}(z_i)  \right]~.
\end{eqnarray}
Then the equivalence can be shown by setting $\eta_t \beta(z_i, t) = d_{\theta_D^{t-1}}(G_{\theta_G^{t-1}}(z_i))~.$

\noindent In practice, the vanilla GAN faces the problem of gradient vanishing on the generator update.  A common trick applied is to use the ``non-saturating loss'', i.e., the generator update is by instead minimizing $-\mathbb{E}_{z\sim \pi(z)}[\log(d_{\theta_D}(G_{\theta_G}(z)))]$. Hence, the gradient of the generator is
\begin{eqnarray}
&& -\nabla_{\theta_G}\log(d_{\theta_D}(G_{\theta_G}(z))) \nonumber\\
&& =   \nabla_{\theta_G}\log(1+e^{-D_{\theta_D}(G_{\theta_G}(z))})\nonumber\\
&& = - \frac{e^{-D_{\theta_D}(G_{\theta_G}(z))}}{1+e^{-D_{\theta_D}(G_{\theta_G}(z))}}\nabla_{\theta_G}D_{\theta_D}(G_{\theta_G}(z))\nonumber\\
&& = -(1- d_{\theta_D}(G_{\theta_G}(z)) )\nabla_{\theta_G}D_{\theta_D}(G_{\theta_G}(z))\nonumber\\
&& = -(1- d_{\theta_D}(G_{\theta_G}(z)) )\nabla D_{\theta_D}(G_{\theta_G}(z))\cdot \nabla_{\theta_G}G_{\theta_G}(z)~.
\end{eqnarray}

\noindent Similarly, with the discriminator parameter $\theta_D^{t-1}$, we can update generator by the following equation
\begin{eqnarray}
&& \theta_G^{t} = \theta_G^{t-1}  + \lambda_t \left[\frac{1}{n}\sum_{i=1}^{n}(1-d_{\theta_D^{t-1}}(G_{\theta_G^{t-1}}(z_i))) \nabla D_{\theta_D^{t-1}}(G_{\theta_G^t}(z_i))\cdot\nabla_{\theta_G}G_{\theta_G^{t-1}}(z_i)  \right]
\end{eqnarray}
Then the equivalence can be shown by setting $\eta_t \beta(z_i, t) = 1-d_{\theta_D^{t-1}}(G_{\theta_G^{t-1}}(z_i))~.$
\begin{remark}
The DiffFlow-GANs formulation provides a more intuitive explanation on why non-saturating loss can avoid vanishing gradients for a poor-trained generator: if at time $t-1$, we have a poor generator $G_{\theta_G^{t-1}}(z)$, generating poor samples that are far from the real data; then $d_{\theta_D^{t-1}}(G_{\theta_G^{t-1}}(z_i))$ would close to $0$, which would lead to zero particle update gradient for original GANs while the ``non-saturating'' loss can avoid this problem. 

In practice, one can avoid the gradient vanishing for DiffFLow-GANs by two methods: either by setting $\beta(z_i, t)\equiv 1$ to maintain the gradient for particle updates; or proposing a noising annealing strategy for the discriminator: during the early stage of training, the discriminator is weakened by classifying a noise-corrupted target distribution $q(x; \sigma(t))$ from fake data $p_t(x)$. The weakening discriminator trick has been adopted in many real deployed GAN models, and it has been shown helpful during the early stage of GAN training \cite{salimans2016improved}. Since the noise annealing on discriminator shares some spirits with SDMs, we will discuss in details later. 
\end{remark}

\subsubsection{Three Improvements on Vanilla GANs}
From previous analysis, we show that the DiffFlow framework provides a novel view on GANs and has potential for several improvements on the vanilla GANs algorithm. The learning dynamics of vanilla GANs are coarse approximation of DiffODE: the one-step gradient of the generator is determined by the particle movements driven by DiffODE, and the driven force is exactly the gradient field of the logistic classifier between the real data and fake data (i.e., discriminator). Furthermore, for the vanilla GAN, the particle gradient is scaled by the probability of the particle being real data, which would be near zero at the early stage of training --- this is exactly the source of gradient vanishing. From this perspective, we can obtain the following improvements:  simplify $\beta(t, X_t)\equiv \beta(t)$, i.e., eliminating the dependence of the particle movement on the scaling factor that determined by the probability of the particle being real.  This would alleviate the gradient vanishing at the early stage of training. 

\noindent Furthermore, since the vanilla GAN only approximates the particle movements by one-step gradients of the least square regression, the is too coarse to simulate the real dynamics of DiffODE. Indeed, one could directly composite a generator by the gradient fields of discriminators at each time step $t$ and this generator could directly simulate the original particle movements in the DiffODE. The idea can be implemented by borrowing ideas from diffusion models: we adopt a time-indexed neural network discriminator $d_{\theta_t}(x, t)$ that is trained by classifying real and fake data at time $t$. 

\noindent Lastly, since at the early stage of training, the generator could face too much pressure with a ``smart'' discriminator, the transition and training dynamics between noise to data could be sharp and unstable during the early stage of training. To achieve a smooth the transition between noise to data, we borrow again ideas from diffusion models: we adopt a the noise annealing strategy $\sigma(t)$ that weakening the discriminator. At time $t$, the discriminator $d_{\theta_t}(x, t)$ learns to classify between noise-corrupted real data $q(x;\sigma(t))$ and fake data $p_t(x)$ where the corruption $\sigma(t)$ is continuously decreasing as time index increasing with $\sigma(0)= \sigma_{max}$ and $\sigma(T)= \sigma_{min}$. This idea is analogous to diffusion models such as NCSN \cite{song2019generative} and the only difference is that the diffusion models learn the score of noise corrupted target distribution $q(x;\sigma(t))$ instead of a classifier at the time index $t$. 

\noindent With above three improvements inspired from the perspective of ODE approximations and diffusion models, we propose an improved GAN algorithm. The training and sampling procedure is described as follows. 

\begin{algorithm} [H]
        \caption{Improved-DiffFlow-GANs-Training}
        \label{alg4}
        \textbf{INPUT: target data from $q(x)$: $x_1^*,\ldots,x_n^*\in\mathbb{R}^k$; noise annealing strategy $\{\sigma_i\}_{i=1}^{T}$; noisy samples $x_0^0, \ldots, x_n^0\in\mathbb{R}^k$ generated from noisy distribution $\pi(x)$; meta-parameter: $T$; time-indexed neural network classifier: $D_{\theta_t}(x, t)$.} \vskip 3mm
        \begin{algorithmic} 
        \For  {$t=1,\ldots, T$} \vskip 1mm
        Sample $z_1, \ldots, z_n$ from $\mathcal{N}(0, \sigma^2_iI)$. \vskip 1mm
        Let $ \theta_t^{*} = \arg\min_{\theta_D} \left[ \frac{1}{n}\sum_{i=1}^{n}\log\left(1+e^{-D_{\theta_t}(x_i^*+z_i, t)}\right) + \frac{1}{n}\sum_{i=1}^{n}\log\left(1+e^{D_{\theta_t}(x_i^{t-1}, t)}\right)   \right]$ \vskip 1mm
       Update the sampled particles: $x_i^{t} = x_i^{t-1} + \eta_t\beta(t)\nabla_{\theta_t} D_{\theta_t^{*}}(x_i^{t-1}, t)$ for  $i=1,\ldots, n$
         \EndFor\\
         \Return Particles $\{x_i^{T}\}_{i=1}^n~$ and $\{\theta_t^*\}_{t=1}^{T}$.
        \end{algorithmic}
\end{algorithm}

\begin{algorithm} [H]
        \caption{Improved-DiffFlow-GANs-Sampling}
        \label{alg4}
        \textbf{INPUT: Noisy distribution $\pi(x)$;  time-indexed discriminator: $D_{\theta_t^*}(x, t)$ for $t=1, \ldots, T$.} \vskip 3mm
    \quad    Sample $X_0\sim \pi(x)$ \vskip 1mm
        \begin{algorithmic} 
        \For  {$t=1,\ldots, T$} \vskip 1mm
       Update the sampled particles: $X_T = X_{T-1} + \eta_t\beta(t)\nabla_{\theta_t} D_{\theta_t^{*}}(x_{T-1}, T-1)$
         \EndFor\\
         \Return $X_T$.
        \end{algorithmic}
\end{algorithm}

\noindent Although we adopt the method of training discriminators $\theta_t^*$ independently across time steps in the algorithm's pseudocode, since it avoids the slow convergence that is partly due to conflicting optimization directions between different time steps \cite{hang2023efficient}. It worths mentioning that our framework offers much more flexibility on designing the time-indexed discriminator: we can either share a universal $\theta$ across all time $t$ as done in diffusion models, or train discriminators $\theta_t^*$ independently for each $t$.  It remains an open problem on which method is better for such generative models.

\subsection{A Unified SDE Framework}
\begin{figure}[H] 
\centering
\includegraphics[width=8cm]{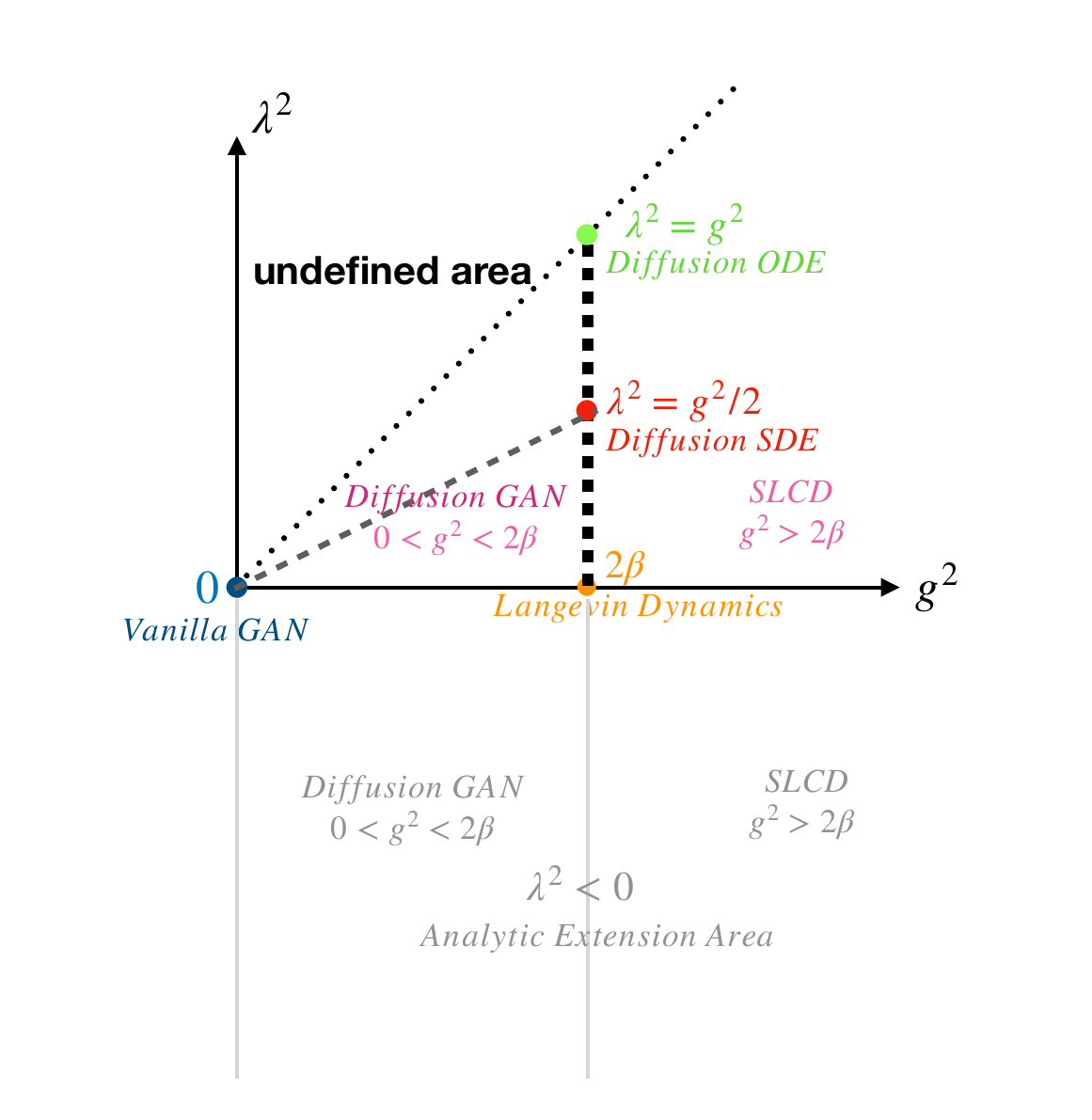} 
\caption{A Continuous Spectrum of Generative Models (Best Viewed in Color).} \label{SPEC}
\end{figure}

In previous sections, we have shown that by specializing scaling functions of DiffFlow, we can recover dynamics of the single-noise Langevin algorithm, diffusion models, and GANs. In fact, DiffFlow offers a much wider continuous spectrum of generative models where GANs and SDMs are corner-cases on the spectrum, as shown in Figure \ref{SPEC}.

\subsubsection{DiffFlow Decomposition}
Recall that the DiffFlow dynamics can be described as the following SDE\footnote{For the ease of presentation, we remove the dependence of $\beta(t, X_t)$ on $X_t$.}:
\begin{eqnarray} 
dX_t =\left[\underbrace{f(X_t, t)}_{Regularization} +\underbrace{ \beta(t)\nabla\log \frac{q(u(t)X_t; \sigma(t))}{p_t(X_t)}}_{Discriminator}+ \underbrace{ \frac{g^2(t)}{2}\nabla\log p_t(X_t)}_{Denoising}\right]dt  + \underbrace{ \sqrt{g^2(t)-\lambda^2(t)}dW_t}_{Diffusion}~. \nonumber
\end{eqnarray}
We name each term in DiffFlow SDE according to its physical meaning to the particle evolutions: the first term $f(X_t, t)$ is usually a linear function of $X_t$ that acts similarly to weight decay for particle updates; the second term is the gradient of the classifier between target data and real data, hence it is named discriminator; the third term is named denoising, since it removes gaussian noise of standard deviation $g(t)$ according to the Kolmogorov Forward Equation; the last term is obviously the diffusion. 

\noindent While the above explanation is obvious in physical meanings, it it hard to explain the continuous evolution of models between GANs and SDMs by changing the scaling functions $g(\cdot)$ and $\lambda(\cdot)$. Notice that when $g^2(t)\leq \beta(t)$, DiffFlow can also be written as follows:
 \begin{eqnarray} 
dX_t =\underbrace{\left[\frac{g^2(t)}{2}\nabla\log q(u(t)X_t; \sigma(t))\right]dt + \sqrt{g^2(t)-\lambda^2(t)}dW_t}_{SDMs}  + \underbrace{\left[\left(\beta(t)-\frac{g^2(t)}{2}\right)\nabla\log \frac{q(u(t)X_t; \sigma(t))}{p_t(X_t)}+f(X_t, t) \right]}_{Regularized\text{ }\text{ }  GANs}dt~. 
\end{eqnarray}
The above decomposition implies that when $g^2(t)\leq 2\beta(t)$, DiffFlow can be seen as a mixed particle dynamics between GANs and SDMs. The relative mixture weight is controlled by $g(t)$ and $\beta(t)$. If we fix $\beta(t)$, then the GANs dynamics would vanish to zero as $g(t)$ increase to $\sqrt{2\beta(t)}$. In the limit of $g(t)=\sqrt{2\beta(t)}$ and $\lambda(t)\equiv 0$, DiffFlow reduces to the pure Langevin algorithm; in the limit of $g(t)\equiv 0$ and $\lambda(t)\equiv 0$, DiffFlow reduces to the pure GANs algorithm. 

\noindent As for the evolution from the pure Langevin algorithm to the diffusion SDE models, we need to increase $\lambda(t)$ from $0$ to $g(t)/\sqrt{2}$ to match the stochasticity of VP/VE SDE. If we further increase $\lambda(t)$ to $g(t)$, we would obtain the diffusion ODE \cite{songscore}. 

\subsubsection{Stochastic Langevin Churn Dynamics (SLCD)}
\noindent If $g(\cdot)$ to $g^2(t)>2\beta(t)$, the GAN component would vanish and we would obtain a Langevin-like algorithm described by the following SDE:
 \begin{eqnarray} 
dX_t = \underbrace{ \left[f(X_t, t) + \beta(t)\nabla\log q(u(t)X_t; \sigma(t))\right]}_{Regularized\text{ }\text{ } Score\text{ }\text{ } Dynamics}dt + \underbrace{  \left( \frac{g^2(t)}{2}- \beta(t)\right)\nabla\log p_t(X_t)}_{Denoising}dt + \underbrace{ \sqrt{g^2(t)-\lambda^2(t)}dW_t}_{Diffusion}~. \nonumber
\end{eqnarray}
We name the above SDE dynamics the Stochastic Langevin Churn Dynamics (SLCD) where we borrow the word ``churn'' from the section $4$ of the EDM paper \cite{karras2022elucidating}, which describes a general Langevin-like process of adding and removing noise according to diffusion and score matching respectively.  The SDE dynamics described above is to some sense exactly the Langevin-like churn procedure in the equation $(7)$ of \cite{karras2022elucidating}: the particle is first raised by a gaussian noise of standard deviation $\sqrt{g^2(t)-\lambda^2(t)}$ and then admits a deterministic noise decay of standard deviation of $ \sqrt{g^2(t)-2\beta(t)}$. This is where the name Stochastic Langevin Churn Dynamics  comes from. 

\subsubsection{Analytic Continuation of DiffFlow}
From previous analysis, we know that the scaling function $g(\cdot)$ controls the proportion of GANs component and $\lambda(\cdot)$ controls the stochasticity and aligns the noise level among Langevin algorithms, diffusion SDEs, and diffusion ODEs. We will show later that the change of $g(t)$ would not affect the marginal distributions $p_t(x)$ and only $\lambda(\cdot)$ plays a critical role in controlling the stochasticity of particle evolution. Further more, we could extend to $\lambda^2(t)<0$ to enable more stochasticity than Langevin algorithms: by letting $\widetilde{\lambda}(t) =\sqrt{-1}\lambda(t) $, we can obtain the following analytic continuation of DiffFlow on $\lambda(t)$:
\begin{eqnarray} 
dX_t =\left[\underbrace{f(X_t, t)}_{Regularization} +\underbrace{ \beta(t)\nabla\log \frac{q(u(t)X_t; \sigma(t))}{p_t(X_t)}}_{Discriminator}+ \underbrace{ \frac{g^2(t)}{2}\nabla\log p_t(X_t)}_{Denoising}\right]dt  + \underbrace{ \sqrt{g^2(t)+\widetilde{\lambda}^2(t)}dW_t}_{Diffusion}~. \nonumber
\end{eqnarray}
As shown in Figure \ref{SPEC}, the analytic continuation area is marked as grey that enables DiffFlow achieves arbitrarily level of stochasticity by controlling $\widetilde{\lambda}(t)$ wisdomly.

\subsubsection{Diffusion-GAN: A Unified Algorithm}
Notice that  when $0<g(t)<\sqrt{2\beta(t)}$, the DiffFlow admits the following mixed particle dynamics of SDMs and GANs that we name Diffusion-GANs, 
\begin{eqnarray} 
dX_t =\underbrace{\left[\frac{g^2(t)}{2}\nabla\log q(u(t)X_t; \sigma(t))\right]dt + \sqrt{g^2(t)-\lambda^2(t)}dW_t}_{SDMs}  + \underbrace{\left[\left(\beta(t)-\frac{g^2(t)}{2}\right)\nabla\log \frac{q(u(t)X_t; \sigma(t))}{p_t(X_t)}+f(X_t, t) \right]}_{Regularized\text{ }\text{ }  GANs}dt~.  
\end{eqnarray}
One can implement Diffusion-GANs under DiffFlow framework by learning a time-indexed discriminator  using logistic regression $$d_{\theta_t^*}(x, t)\approx \log \frac{q(u(t)X_t; \sigma(t))}{p_t(X_t)}~$$ and a score network using score mathcing $$s_{\theta'_*}(x, t)\approx \nabla\log q(u(t)X_t; \sigma(t))~.$$ Then the sampling process is defined by discretizing the following SDE within interval $t\in [0, T]$: 
\begin{eqnarray} 
dX_t =\left[\frac{g^2(t)}{2}s_{\theta'_*}(x, t) \right]dt + \sqrt{g^2(t)-\lambda^2(t)}dW_t  + \left[\left(\beta(t)-\frac{g^2(t)}{2}\right)\nabla d_{\theta_t^*}(x, t) +f(X_t, t) \right]dt~.  
\end{eqnarray}
Notice that we should tune $\lambda(t)$ to achieve a good stochasticity. It remains an open problem empirically and theoretically that how to design an optimal noise level to achieve the best sample generation quality. 

\subsubsection{Marginal Preserving Property}
One may want to argue that the current framework of DiffFlow is just a simple combination of particle dynamics of GANs and SDMs respectively, where $g(\cdot)$ acts as an interpolation weight between them. It is already known in the literature that both GANs \cite{gao2019deep, yi2023monoflow} and SDMs \cite{songscore} can be modelled as particle dynamics with different ordinary or stochastic differential equations.  

\noindent We want to argue that despite both GANs and SDMs can be modelled as ODE/SDE with different drift and diffusion terms, the dynamics behind these differential equations are different: even with the same initializations of particles, the path measure would deviate significantly for GANs and SDMs. Hence, a simple linear interpolation between dynamics of GANs and SDMs lacks both theoretic and practical motivations and we need sophisticated design of each term in the differential equations to align the marginal distribution of different classes of generative models lied in SDMs and GANs. This work provides a reasonable design of a unified SDE ``DiffFlow'' for GANs and SDMs that enables flexible interpolations between them by changing the scaling functions in the SDE. Furthermore, we would show in the following proposition that the interpolation is ``smooth'' from GANs to SDMs: the marginal distribution $p_t(x)$ for all $t\geq 0$ would remain invariant to the interpolation factor $g(\cdot)$. 

\begin{proposition} [Marginal Preserving Property]
The marginal distribution $p_t(x)$ of DiffFlow 
\begin{eqnarray} 
dX_t =\left[f(X_t, t) + \beta(t)\nabla\log \frac{q(u(t)X_t; \sigma(t))}{p_t(X_t)}+ \frac{g^2(t)}{2}\nabla\log p_t(X_t)\right]dt  +  \sqrt{g^2(t)-\lambda^2(t)}dW_t~. \nonumber
\end{eqnarray}
remains invariant to $g(\cdot)$. 
\begin{proof}
By the Kolmogorov Forward Equation \cite{oksendal2003stochastic}, the marginal distribution $p_t(x)$ follows the following PDE:
\begin{eqnarray} 
&&\frac{\partial p_t(x)}{\partial t} \nonumber\\
&& = -\nabla\cdot\left[p_t(x)\left(f(x, t) + \beta(t)\nabla\log \frac{q(u(t)x; \sigma(t))}{p_t(x)}+ \frac{g^2(t)}{2}\nabla\log p_t(x)\right)\right] +\frac{g^2(t)-\lambda^2(t)}{2}\nabla\cdot\nabla p_t(x)\nonumber\\
&& =  -\nabla\cdot\left[p_t(x)\left(f(x, t) + \beta(t)\nabla\log \frac{q(u(t)x; \sigma(t))}{p_t(x)}\right]\right) - \nabla\cdot\left[p_t(x)\frac{g^2(t)}{2}\nabla\log p_t(x)\right]  +\frac{g^2(t)-\lambda^2(t)}{2}\nabla\cdot\nabla p_t(x) \nonumber\\
&&  =  -\nabla\cdot\left[p_t(x)\left(f(x, t) + \beta(t)\nabla\log \frac{q(u(t)x; \sigma(t))}{p_t(x)}\right]\right) -\frac{\lambda^2(t)}{2}\nabla\cdot\nabla p_t(x)  ~.
\end{eqnarray}
Hence, the marginal distribution $p_t(x)$ is independent of $g(\cdot)$. 
\end{proof}
\end{proposition}
\noindent The Marginal Preserving Property implies that if we increase $g(\cdot)$, the DiffFlow evolutes from GANs to SDMs and further to SLCD as shown in the Figure \ref{SPEC}, while the marginal distribution $p_t(x)$ remains invariant.  Another important factor $\lambda(\cdot)$ is used to control the stochasticity of DiffFlow to align the noise level among Langevin Dynamics, diffusion SDEs, and diffusion ODEs.

\section{Convergence Analysis}
For the ease of presentation, without loss of generality, we set $f(X_t, t) \equiv 0$, $\beta(t, X_t)\equiv 1$, $u(t)\equiv 1$, $\sigma(t)\equiv \sigma_0$ for some $\alpha, \sigma_0>0$ and $\lambda(t)\equiv 0$, then the DiffFlow becomes the following SDE:
\begin{eqnarray} 
dX_t =\left[\nabla\log \frac{q(X_t; \sigma_0)}{p_t(X_t)}+  \frac{g^2(t)}{2}\nabla\log p_t(X_t)\right]dt  +  g(t)dW_t~. \nonumber
\end{eqnarray}
\begin{remark}
The current simplified DiffFlow is general enough to incorporate the original DiffFlow's full dynamics: by rescaling the particles and gradients, we can recover general $f(X_t, t)$, $\beta(t)$, and $u(t)$, as done in the DDIM \cite{song2020denoising}. The general $\lambda(t)$ and $\sigma(t)$ can be recovered by changing the noise scheduler from constants to the desired annealing scheduler.
\end{remark}

\noindent In order to study the convergence of DiffFlow dynamics, we need to find its variational formulation, i.e.,  a functional that DiffFlow minimizes. Finding such functional is not hard, as shown in the following lemma, which implies the functional is exactly the KL divergence between the generated and target distributions. 
\begin{lemma}[The Variational Formulation of DiffFlow] \label{VFD}
Given stochastic process $\{X_t\}_{t\geq 0}$ and its dynamics determined by 
\begin{eqnarray} 
dX_t =\left[\nabla\log \frac{q(X_t; \sigma_0)}{p_t(X_t)}+  \frac{g^2(t)}{2}\nabla\log p_t(X_t)\right]dt  +  g(t)dW_t~
\end{eqnarray}
with $X_0\sim \pi(x)$ and $\sigma_0, \lambda_0>0$. Then the marginal distribution $p_t(x)$ of $X_t$ minimizes the following functional
\begin{eqnarray}
&& L(p) = KL( p\| q(x;\sigma_0)) := \int_{\mathbb{R}^k}p(x)\log\frac{p(x)}{q(x;\sigma_0)}dx ~.
\end{eqnarray}
Furthermore, 
\begin{eqnarray}
&&  \frac{\partial L(p_t)}{\partial t}  = -  \int_{\mathbb{R}^k}p_t(x)\left\|\nabla\log \frac{p_t(x)}{q(x; \sigma_0)}\right\|_2^2dx~.
\end{eqnarray}
\begin{proof}
By the Kolmogorov Forward Equation, the marginal distribution $p_t(x)$ follows the following PDE:
\begin{eqnarray} 
&&\frac{\partial p_t(x)}{\partial t} \nonumber\\
&& = -\nabla\cdot\left[p_t(x)\left( \nabla\log \frac{q(x; \sigma_0)}{p_t(x)}+ \frac{g^2(t)}{2}\nabla\log p_t(x)\right)\right] +\frac{g^2(t)}{2}\nabla\cdot\nabla p_t(x)\nonumber\\
&& = -\nabla\cdot\left[p_t(x)\left(\nabla\log \frac{q(x; \sigma_0)}{p_t(x)}+ \frac{g^2(t)}{2}\nabla\log p_t(x)\right)\right] +\frac{g^2(t)}{2}\nabla\cdot\left[p_t(x)\nabla \log p_t(x)\right] \nonumber\\
&& = -\nabla\cdot\left[p_t(x)\left( \nabla\log \frac{q(x; \sigma_0)}{p_t(x)}+ \frac{g^2(t)}{2}\nabla\log p_t(x) -\frac{g^2(t)}{2}\nabla\log p_t(x) \right)\right]\nonumber\\
&& =  -\nabla\cdot\left[p_t(x)\left( \nabla\log \frac{q(x; \sigma_0)}{p_t(x)}\right)\right]~.
\end{eqnarray}
Then, we have
\begin{eqnarray}
&& \frac{\partial L(p_t)}{\partial t}   \nonumber\\
&&  =  \int_{\mathbb{R}^k} \left[\log\frac{p_t(x)}{q(x;\sigma_0)} +1\right] \frac{\partial p_t(x)}{\partial t}dx   \nonumber\\
&& =   \int_{\mathbb{R}^k} \log\frac{p_t(x)}{q(x;\sigma_0)} \frac{\partial p_t(x)}{\partial t}dx \nonumber\\
&& =   -   \int_{\mathbb{R}^k} \log\frac{p_t(x)}{q(x;\sigma_0)} \nabla\cdot\left[p_t(x)\left( \nabla\log \frac{q(x; \sigma_0)}{p_t(x)}\right)\right]dx \nonumber\\
&& =     \int_{\mathbb{R}^k} \log\frac{p_t(x)}{q(x;\sigma_0)} \nabla\cdot\left[p_t(x)\left( \nabla \log\frac{p_t(x)}{q(x;\sigma_0)} \right)\right]dx
\end{eqnarray}
Through integral by parts, we have
\begin{eqnarray}
&& \frac{\partial L(p_t)}{\partial t}   \nonumber\\
&&  =-  \int_{\mathbb{R}^k}p_t(x)\left( \nabla \log\frac{p_t(x)}{q(x;\sigma_0)} \right)\cdot \left( \nabla \log\frac{p_t(x)}{q(x;\sigma_0)} \right) dx \nonumber\\
&&  =-  \int_{\mathbb{R}^k}p_t(x)\left\| \nabla \log\frac{p_t(x)}{q(x;\sigma_0)} \right\|_2^2dx~.
\end{eqnarray}
Hence,  the KL divergence $L(p_t)$ is decreasing along the marginal distribution path $\{p_t(x)\}_{t\geq 0}$ determined by DiffFlow.
\end{proof}
\end{lemma}

\subsection{Asymptotic Optimality via Poincare Inequality}
The main tool to prove the asymptotic convergence of DiffFlow is the following Gaussian Poincare inequality from \cite{ledoux2006concentration}. 
\begin{definition}[Gaussian Poincare Inequality]
Suppose $f: \mathbb{R}^k\to\mathbb{R}$ is a smooth function and $X$ is a multivariate Gaussian distribution $X\sim \mathcal{N}(0, \sigma^{2}I)$ where $I\in\mathbb{R}^{k\times k}$. Then
\begin{eqnarray}
Var[f(X)] \leq \sigma^{2} \mathbb{E}\left[\left\|\nabla f(X)\right\|_2^2\right]~.
\end{eqnarray}
\end{definition}

\noindent We also show that under gaussian smoothing, the log-density obeys quadratic growth. 
\begin{lemma}[Bound on the log-density of Smoothed Distributions] \label{lemma1}
Given a probability distribution $q(x)$ and let $q(x;\sigma)$ be the distribution of $x+\epsilon$ where $x\sim q(x)$ and $\epsilon\sim \mathcal{N}(0, \sigma^2 I)$. Then there exists some constants $A_\sigma, B_\sigma>0$ and $C_\sigma$ such that for any $0<\gamma<1$
 \begin{eqnarray}
 \left|\log q(x;\sigma)\right|\leq A_\sigma\|x\|_2^2+B_\sigma\|x\|_2+C_\sigma
 \end{eqnarray}
 where $C_q(\gamma) :=\inf\left\{s : \int_{B(s)}q(u)du\geq \gamma\right\}, A_\sigma =\frac{1}{2\sigma^2}, B_\sigma=\frac{C_p(\gamma)}{\sigma^2}$ and \\ $C_\sigma=\max\left\{\frac{C^2_p(\gamma)}{2\sigma^2}- \log\left(\gamma \frac{1}{(2\pi)^{k/2}\sigma^k}\right), \log\left( \frac{1}{(2\pi)^{k/2}\sigma^k}\right) \right\}$~, $B(s)=\{x\in\mathbb{R}^k: \|x\|_2\leq s\}$~.
 
\begin{proof}
Let $G_\sigma(x)$ be the probability density function of $ N(0, \sigma^2 I)$, then the resulting smoothed distribution $q(x;\sigma)$ is
\begin{eqnarray}
&& q(x;\sigma) = \int_{\mathbb{R}^d}q(u)G_\sigma(x-u)du~.
\end{eqnarray} 
Let $B(r)=\{u: \|u\|_2\leq r\}$, then
\begin{eqnarray}
&&q(x;\sigma) =\int_{B(r)}q(u)G_\sigma(x-u)du + \int_{\mathbb{R}^k\setminus B(r)}q(u)G_\sigma(x-u)du\nonumber\\
&&\geq \int_{B(r)}q(u)G_\sigma(x-u)du \nonumber\\
&&\geq \int_{B(r)}q(u)G_\sigma\left(x+r\frac{x}{\|x\|_2}\right)du \nonumber\\
&&=G_\sigma\left(x+r\frac{x}{\|x\|_2}\right)\int_{B(r)}q(u)du~.
\end{eqnarray} 
Fix some small constant $0<\gamma<1$, if we choose $r= C_p(\gamma) :=\inf\left\{s : \int_{B(s)}p(u)du\geq \gamma\right\}$. This implies
\begin{eqnarray}\label{eqn2.4}
&&q(x;\sigma)  \geq \gamma G_\sigma\left(x+C_p(\gamma)\frac{x}{\|x\|_2}\right) \nonumber\\
&&=\gamma \frac{1}{(2\pi)^{k/2}\sigma^k}\exp\left(\frac{-\left\| x+C_p(\gamma)\frac{x}{\|x\|_2}\right \|_2^2}{2\sigma^2}\right)~,
\end{eqnarray} 
Taking logarithm on both sides of (\ref{eqn2.4}), we obtain
 \begin{eqnarray}
 &&\log q(x;\sigma)\geq \log\left(\gamma \frac{1}{(2\pi)^{k/2}\sigma^k}\right)-\frac{\left\| x+C_p(\gamma)\frac{x}{\|x\|_2}\right \|_2^2}{2\sigma^2}  \nonumber\\
&&= -\frac{1}{2\sigma^2} \|x\|_2^2 -\frac{C_p(\gamma)}{\sigma^2}\|x\|_2-\frac{C^2_p(\gamma)}{2\sigma^2}+ \log\left(\gamma \frac{1}{(2\pi)^{k/2}\sigma^k}\right)~.
 \end{eqnarray} 

\noindent We also have
\begin{eqnarray}
&& q(x;\sigma) = \int_{\mathbb{R}^k}q(u)G_\sigma(x-u)du \nonumber\\
&&\leq  \int_{\mathbb{R}^k}q(u)G_\sigma(0)du  \nonumber\\
&& = G_\sigma(0)  \nonumber\\
&& =  \frac{1}{(2\pi)^{k/2}\sigma^k}~.
\end{eqnarray} 
Therefore, 
\begin{eqnarray}
\log q(x;\sigma)\leq \log\left( \frac{1}{(2\pi)^{k/2}\sigma^k}\right).
\end{eqnarray} 
 Let $A_\sigma =\frac{1}{2\sigma^2}, B_\sigma=\frac{C_p(\gamma)}{\sigma^2}$ and $C_\sigma=\max\left\{\frac{C^2_p(\gamma)}{2\sigma^2}- \log\left(\gamma \frac{1}{(2\pi)^{k/2}\sigma^k}\right), \log\left( \frac{1}{(2\pi)^{k/2}\sigma^k}\right) \right\}$, then
 \begin{eqnarray}
 \left|\log q(x;\sigma)\right|\leq A_\sigma\|x\|_2^2+B_\sigma\|x\|_2+C_\sigma~.
 \end{eqnarray}
 \end{proof}
\end{lemma}
\begin{remark}
In the above Lemma, for the measure $q$, we define the quantity $$C_q(\gamma) :=\inf\left\{s : \int_{B(s)}q(u)du\geq \gamma\right\}~.$$ The meaning of this quantity $C_p(\gamma)$ is the smallest ball centred at the origin of $\mathbb{R}^k$ that captures at least $\gamma$ mass of the probability measure $q$.
\end{remark}
\noindent Now, we are ready to prove the asymptotic convergence theorem.
\begin{theorem}
Given stochastic process $\{X_t\}_{t\geq 0}$ and its dynamics determined by 
\begin{eqnarray} 
dX_t =\left[\nabla\log \frac{q(X_t; \sigma_0)}{p_t(X_t)}+  \frac{g^2(t)}{2}\nabla\log p_t(X_t)\right]dt  +  g(t)dW_t~
\end{eqnarray}
with $X_0\sim \pi(x)$ and $\sigma_0, \lambda_0>0$. Then the marginal distribution $X_t\sim p_t(x)$ converges almost everywhere to the target distribution $q(x;\sigma_0)$, i.e.,
\begin{eqnarray}
\lim_{t\to\infty}p_t(x) = q(x;\sigma_0)\quad \text{a.e.}
\end{eqnarray}
\begin{proof}
By Lemma \ref{VFD}, 
\begin{eqnarray}
  \frac{\partial L(p_t)}{\partial t}  =-  \int_{\mathbb{R}^k}p_t(x)\left\| \nabla \log\frac{p_t(x)}{q(x;\sigma_0)} \right\|_2^2dx.
\end{eqnarray}
Hence, by the nonnegativity of KL divergence, 
\begin{eqnarray}
\lim_{t\to\infty} \int_{\mathbb{R}^k}p_t(x)\left\| \nabla \log\frac{p_t(x)}{q(x;\sigma_0)} \right\|_2^2dx = 0~. 
\end{eqnarray}
Furthermore, since
\begin{eqnarray}
&&\left\|\nabla \sqrt{f(x)}\right\|_2^2 = \frac{\left\|\nabla f(x)\right\|_2^2}{4f(x)} = \frac{f(x)}{4} \left\|\nabla \log f(x)\right\|_2^2 
\end{eqnarray} 
we have 
\begin{eqnarray}
&&\int_{\mathbb{R}^k}p_t(x)\left\|\nabla\log \frac{p_t(x)}{q(x; \sigma_0)}\right\|_2^2dx\nonumber\\
&& = 4\int_{\mathbb{R}^k}q(x;\sigma_0)\left\|\nabla\sqrt{ \frac{p_t(x)}{q(x; \sigma_0)}}\right\|_2^2dx   \nonumber\\
&& = 4\int_{\mathbb{R}^k}\exp\left(\log q(x;\sigma_0)\right)\left\|\nabla\sqrt{ \frac{p_t(x)}{q(x; \sigma_0)}}\right\|_2^2\nonumber\\
&& \geq 4\int_{\mathbb{R}^k}\exp\left(-A_{\sigma_0}\|x\|_2^2-B_{\sigma_0}\|x\|_2-C_{\sigma_0})\right)\left\|\nabla\sqrt{ \frac{p_t(x)}{q(x; \sigma_0)}}\right\|_2^2dx ~.
\end{eqnarray} 
Since $\|x\|_2\leq \|x\|_2^2+1$, we have 
\begin{eqnarray}
&&\int_{\mathbb{R}^k}p_t(x)\left\|\nabla\log \frac{p_t(x)}{q(x; \sigma_0)}\right\|_2^2dx\nonumber\\
&& \geq 4\int_{\mathbb{R}^k}\exp\left(-\left(A_{\sigma_0}+B_{\sigma_0}\right)\|x\|_2^2-C_{\sigma_0}-1\right)\left\|\nabla\sqrt{ \frac{p_t(x)}{q(x; \sigma_0)}}\right\|_2^2dx \nonumber\\
&&=  4\exp\left(-C_{\sigma_0}-1\right)\int_{\mathbb{R}^k}\exp\left(-\frac{\|x\|_2^2}{\left(A_{\sigma_0}+B_{\sigma_0}\right)^{-1}}\right)  \left\|\nabla\sqrt{ \frac{p_t(x)}{q(x; \sigma_0)}}\right\|_2^2dx \nonumber\\
&&=  4\left(\frac{\pi}{A_{\sigma_0}+B_{\sigma_0}}\right)^{k/2}\exp\left(-C_{\sigma_0}-1\right)\int_{\mathbb{R}^k}\mathcal{N}\left(x; 0, \frac{1}{2(A_{\sigma_0}+B_{\sigma_0})}I\right)\left\|\nabla\sqrt{ \frac{p_t(x)}{q(x; \sigma_0)}}\right\|_2^2dx \nonumber\\
&&=4\left(\frac{\pi}{A_{\sigma_0}+B_{\sigma_0}}\right)^{k/2}\exp\left(-C_{\sigma_0}-1\right)\mathbb{E}_{x\sim\mathcal{N}\left(x; 0, \frac{1}{2(A_{\sigma_0}+B_{\sigma_0})}I\right) }\left(\left\|\nabla\sqrt{ \frac{p_t(x)}{q(x; \sigma_0)}}\right\|_2^2\right) \nonumber\\
&&\geq 8\left(A_{\sigma_0}+B_{\sigma_0}\right)\left(\frac{\pi}{A_{\sigma_0}+B_{\sigma_0}}\right)^{k/2}\exp\left(-C_{\sigma_0}-1\right)Var_{x\sim\mathcal{N}\left(x; 0, \frac{1}{2(A_{\sigma_0}+B_{\sigma_0})}I\right) }\left( \sqrt{ \frac{p_t(x)}{q(x; \sigma_0)}}\right) ~.
\end{eqnarray} 
where the last inequality is due to Gaussian Poincare inequality. Hence, we obtain
\begin{eqnarray}
\lim_{t\to\infty}Var_{x\sim\mathcal{N}\left(x; 0, \frac{1}{2(A_{\sigma_0}+B_{\sigma_0})}I\right) }\left(  \sqrt{ \frac{p_t(x)}{q(x; \sigma_0)}}\right) = 0~.
\end{eqnarray}
Furthermore, by previous analysis on the lower bound of $\exp\left(\log q(x;\sigma_0)\right)$, we have
\begin{eqnarray}
\exp\left(\log q(x;\sigma_0)\right)\geq  D_{\sigma_0}\mathcal{N}\left(x; 0, \frac{1}{2(A_{\sigma_0}+B_{\sigma_0})}I\right)~.
\end{eqnarray}
where  $D_{\sigma_0}:=4\left(\frac{\pi}{A_{\sigma_0}+B_{\sigma_0}}\right)^{k/2}\exp\left(-C_{\sigma_0}-1\right)$~. 

\noindent Then
\begin{eqnarray}
 && \mathbb{E}_{x\sim\mathcal{N}\left(x; 0, \frac{1}{2(A_{\sigma_0}+B_{\sigma_0})}I\right) }\left(\sqrt{ \frac{p_t(x)}{q(x; \sigma_0)}}\right)\nonumber\\
 && = \int_{\mathbb{R}^k}\mathcal{N}\left(x; 0, \frac{1}{2(A_{\sigma_0}+B_{\sigma_0})}I\right)\sqrt{ \frac{p_t(x)}{q(x; \sigma_0)}}dx\nonumber\\
 && \leq \frac{1}{\sqrt{D_{\sigma_0}}} \int_{\mathbb{R}^k}\sqrt{\mathcal{N}\left(x; 0, \frac{1}{2(A_{\sigma_0}+B_{\sigma_0})}I\right)}\sqrt{p_t(x)}dx\nonumber\\
 && \leq  \frac{1}{\sqrt{D_{\sigma_0}}} \int_{\mathbb{R}^k}\mathcal{N}\left(x; 0, \frac{1}{2(A_{\sigma_0}+B_{\sigma_0})}I\right)dx\int_{\mathbb{R}^k}p_t(x)dx\nonumber\\
 && \leq \frac{1}{\sqrt{D_{\sigma_0}}}~.
\end{eqnarray}

Hence,
\begin{eqnarray}
\lim_{t\to\infty} \sqrt{ \frac{p_t(x)}{q(x; \sigma_0)}}  = const. \leq \frac{1}{\sqrt{D_{\sigma_0}}}<\infty
\end{eqnarray}

This implies
\begin{eqnarray}
\lim_{t\to\infty}p_t(x) = q(x;\sigma_0)\quad \text{a.e.}
\end{eqnarray}
\end{proof}
\end{theorem}

\subsection{Nonasymptotic Convergence Rate via Log-Sobolev Inequality}
In the previous section, we have proved the asymptotic optimality of DiffFlow. In order to obtain an explicit convergence rate to the target distribution, we need stronger functional inequalities, i.e., the log-Sobolev inequality \cite{ledoux2006concentration, ma2019sampling}.
\begin{definition}[Log-Sobolev Inequality]
For a smooth function $g: \mathbb{R}^k\to\mathbb{R}$, consider the Sobolev space defined by the weighted $L^2$ norm: $\|g\|_{L^2(q)} = \int_{\mathbb{R}^k} g(x)^2q(x)dx$. We say $q(x)$ satisfies the log-Sobolev inequality with constant $\rho>0$ if the following inequality holds for any $\int_{\mathbb{R}^k}g(x)q(x)=1$, 
\begin{eqnarray}
&&  \int_{\mathbb{R}^k} g(x)\log g(x)\cdot q(x)dx\leq \frac{2}{\rho}\int_{\mathbb{R}^k}\left\|\nabla\sqrt{g(x)}\right\|_2^2q(x)dx~.
\end{eqnarray}
\end{definition}
\noindent Then we can obtain a linear convergence of marginal distribution $p_t(x)$ to the smoothed target distribution $q(x;\sigma_0)$. The analysis is essentially the same as Langevin dynamics as in \cite{ma2019sampling}, since their marginal distribution shares the same Fokker-Planck equation. 
\begin{theorem}
Given stochastic process $\{X_t\}_{t\geq 0}$ and its dynamics determined by 
\begin{eqnarray} 
dX_t =\left[\nabla\log \frac{q(X_t; \sigma_0)}{p_t(X_t)}+  \frac{g^2(t)}{2}\nabla\log p_t(X_t)\right]dt  +  g(t)dW_t~
\end{eqnarray}
with $X_0\sim \pi(x)$ and $\sigma_0, \lambda_0>0$.  If the smoothed target distribution $q(x;\sigma_0)$  satisfies the log-Sobolev inequality with constant $\rho_q>0$. Then the marginal distribution $X_t\sim p_t(x)$ converges linear to the target distribution $q(x;\sigma_0)$ in KL divergence, i.e.,
\begin{eqnarray}
KL(p_t(x)\|q(x;\sigma_0)) \leq \exp(-2\rho_q t) KL(\pi(x)\|q(x;\sigma_0)).
\end{eqnarray}
\begin{proof}
From Lemma \ref{VFD}, we obtain
\begin{eqnarray}
&&  \frac{\partial L(p_t)}{\partial t}=\frac{\partial }{\partial t}KL(p_t(x)\|q(x;\sigma_0))  = -  \int_{\mathbb{R}^k}p_t(x)\left\|\nabla\log \frac{p_t(x)}{q(x; \sigma_0)}\right\|_2^2dx~.
\end{eqnarray}
Since $q(x;\sigma_0)$  satisfies the log-Sobolev inequality with constant $\rho_q>0$, by letting the test function $g(x) = p_t(x)/q(x;\sigma_0)$, we obtain
\begin{eqnarray}
&& KL(p_t(x)\|q(x;\sigma_0))\nonumber\\
&&   \leq \frac{2}{\rho_q} \int_{\mathbb{R}^k}\left\|\nabla\sqrt{\frac{p_t(x)}{q(x;\sigma_0)}}\right\|_2^2q(x;\sigma_0)dx~.
\end{eqnarray}
Since we already know from previous analysis that $$\left\|\nabla \sqrt{f(x)}\right\|_2^2 = \frac{f(x)}{4} \left\|\nabla \log f(x)\right\|_2^2~$$
we have
\begin{eqnarray}
&& KL(p_t(x)\|q(x;\sigma_0))\nonumber\\
&&   \leq \frac{1}{2\rho_q} \int_{\mathbb{R}^k}p_t(x)\left\|\nabla\frac{p_t(x)}{q(x;\sigma_0)}\right\|_2^2q(x)dx\nonumber\\
&&  =  - \frac{1}{2\rho_q} \frac{\partial }{\partial t}KL(p_t(x)\|q(x;\sigma_0))~.
\end{eqnarray}
Hence by Grönwall's inequality, 
\begin{eqnarray}
KL(p_t(x)\|q(x;\sigma_0)) \leq \exp(-2\rho_q t) KL(\pi(x)\|q(x;\sigma_0)).
\end{eqnarray}
\end{proof}
\end{theorem}

\noindent Different from asymptotic analysis in the previous subsection that holds for general $q(x)$, the convergence rate is obtained under the assumption that the smoothed target $q(x;\sigma_0)$ satisfies the log-Sobolev inequalities.  This is rather a strong assumption, since we need control the curvature lower bound of smoothed energy function $U(x; q; \sigma_0)$ to satisfy the Lyapunov conditions for log-Sobolev inequality \cite{cattiaux2010note}. 
This condition holds for some simple distribution: if $q(x)\sim \mathcal{N}(\mu_q, \sigma_q^2I)$, then
the Hessian of its smoothed energy function $U(x; q; \sigma_0)=-\nabla^2\log q(x;\sigma_0) = (\sigma_0^2+\sigma_q^2)^{-1}I$, then by Bakery-Emery criteria \cite{bakry2006diffusions}, we have $\rho_q = (\sigma_0^2+\sigma_q^2)^{-1}$~.  However, for a general target $q(x)$, obtaining the log-Sobolev inequality is relatively hard. If we still want to obtain an explicit convergence rate, we can seek for an explicit regularization $f(t, X_t)$ on the DiffFlow dynamics to restrict the path measure in a smaller subset and then the explicit convergence rate can be obtained by employing uniform log-Sobolev inequalities along the path measure \cite{guillin2022uniform}. The detailed derivation of such log-Sobolev inequalities under particular regularization $f(t, X_t)$ on the path measure is beyond the scope of this paper and we leave it as future work.

\section{Maximal Likelihood Inference}
Recall that the dynamics of DiffFlow is described by the following SDE on the interval $t\in [0, T]$ with $X_0\sim\pi(x)$ and $X_t\sim p_t(x)$: 
\begin{eqnarray} 
dX_t =\left[f(X_t, t) + \beta(t, X_t)\nabla\log \frac{q(u(t)X_t; \sigma(t))}{p_t(X_t)}+ \frac{g^2(t)}{2}\nabla\log p_t(X_t)\right]dt  + \sqrt{g^2(t)-\lambda^2(t)}dW_t~.
\end{eqnarray}
At the end time step $T$, we obtain $p_T(x)$ and we hope this is a ``good'' approximation to the target distribution. Previous analysis shows that as $T$ to infinity, $p_T(x)$ would converge to the target distribution $q(x;\sigma_0)$ under simplified dynamics. However, the convergence rate is relatively hard to obtain for general target distribution with no isoperimetry property. In this section, we define the goodness of approximation of $p_T(x)$ from another perspective: maximizing the likelihood at the end of time. 

\noindent There are already many existing works on analyzing the likelihood dynamics of diffusion SDE, for instance \cite{huang2021variational, song2021maximum, kingma2021variational}. These works provide a continuous version of ELBO for diffusion SDEs by different techniques. In this section, we follow the analysis of \cite{huang2021variational} that adopts a Feymann-Kac representation on $p_T$ then using the Girsanov change of measure theorem to obtain a trainable ELBO.

 \noindent Here we mainly consider when $g^2(t)\leq \beta(t)$, DiffFlow can also be seen as a mixed particle dynamics between SDMs and GANs. Recall in this case, DiffFlow can be rewritten as, 
 \begin{eqnarray} 
dX_t =\underbrace{\left[\frac{g^2(t)}{2}\nabla\log q(u(t)X_t; \sigma(t))\right]dt + \sqrt{g^2(t)-\lambda^2(t)}dW_t}_{SDMs}  + \underbrace{\left[\left(\beta(t)-\frac{g^2(t)}{2}\right)\nabla\log \frac{q(u(t)X_t; \sigma(t))}{p_t(X_t)}+f(X_t, t) \right]}_{Regularized\text{ }\text{ }  GANs}dt~.  \nonumber
\end{eqnarray}

\noindent Given a time-indexed discriminator $d_{\theta_D^t}(x, t): \mathbb{R}^d\times [0, T]\to\mathbb{R}^d$ using logistic regression $$d_{\theta_D^t}(x, t)\approx \log \frac{q(u(t)X_t; \sigma(t))}{p_t(X_t)}~$$ and a score network $s_{\theta_c^t}(x, t): \mathbb{R}^d\times [0, T]\to\mathbb{R}^d$ using score mathcing $$s_{\theta_c^t}(x, t)\approx \nabla\log q(u(t)X_t; \sigma(t))~.$$ Then the approximated process is given by the following neural SDE:
\begin{eqnarray}  \label{NSDE}
dX_t =\left[\frac{g^2(t)}{2}s_{\theta_c^t}(X_t, t) \right]dt + \sqrt{g^2(t)-\lambda^2(t)}dW_t  + \left[\left(\beta(t)-\frac{g^2(t)}{2}\right)\nabla d_{\theta_D^t}(X_t, t) +f(X_t, t) \right]dt~.  
\end{eqnarray}
We need to answer the question of how to train the score networks $s_{\theta_c^t}(x, t)$ and the time-indexed discriminator $d_{\theta_D^t}(x, t)$ that optimizes the ELBO of the likelihood $\log p_T(x)$.

\noindent Following the analysis of \cite{huang2021variational}, in order to obtain a general connection between maximal likelihood estimation and neural network training, we need to apply the Girsanov formula to obtain a trainable ELBO for the likelihood of the terminal marginal density. Before we introduce our main theorem, we need the following two well-known results from stochastic calculus. The first one is Feymann-Kac Formula, adapted from Theorem $7.6$ in \cite{karatzas1991brownian}. 
\begin{lemma}[Feymann-Kac Formula]
Suppose $u(t, x): [0, T]\times\mathbb{R}^d\to \mathbb{R}$ is of class $C^{1, 2}([0, T]\times\mathbb{R}^d])$ and satisfies the following PDE:
\begin{eqnarray}
&&\frac{\partial u(t, x)}{\partial t} + c(x, t)u(t, x) + \frac{\sigma(t)^2}{2}\nabla\cdot\nabla u(t, x)+b(t, x)\cdot \nabla u(t, x) = 0
\end{eqnarray}
with terminal condition $u(T, x) = u_T(x)$. If $u(t, x)$ satisfies the polynomial growth condition
\begin{eqnarray}
\max_{0\leq t\leq T}|u(t, x)| \leq M(1+\|x\|^{2\mu}), \quad x\in\mathbb{R}^d
\end{eqnarray}
for some $M>0$ and $\mu\geq 1$. Then $u(t, x)$ admits the following stochastic representation
\begin{eqnarray}
u(t, x) = \mathbb{E}\left[  u_T(X_T)\exp\left(\int_t^T c(X_s, s) ds\right)   \Bigg| X_t = x\right]
\end{eqnarray}
where $\{X_s\}_{t\leq s\leq T}$ solves the following SDE with initial $X_t = x$,
\begin{eqnarray}
dX_s = b(t, X_s)dt + \sigma(t) dW_t~.
\end{eqnarray}
\end{lemma}
\noindent Then we need the well-known Girsanov Theorem to measure the deviation of path measures.
\begin{lemma}[Girsanov Formula, Theorem $8.6.3$ in \cite{oksendal2013stochastic}]
Let $(\Omega,\mathcal{F},\mathbb{P})$ be the underlying probability space for which $W_s$ is a Brownian motion. Let $\widetilde{W}_s$ be an ito process solving
\begin{eqnarray}
&& d\widetilde{W}_s = a(\omega, s)ds + dW_s
\end{eqnarray}
for $\omega\in\Omega$, $0\leq s\leq T$ and $\widetilde{W}_0=0$~ and $a(\omega, s)$ satisfies the Novikov's condition, i.e., $$\mathbb{E}\left[\exp\left( \frac{1}{2}\int_0^Ta^2(\omega, s)ds\right)\right]<\infty~.$$ Then $\widetilde{W}_s$ is a Brownian motion w.r.t. $\mathbb{Q}$ determined by
\begin{eqnarray}
\log\frac{d\mathbb{P} }{d\mathbb{Q} }(\omega) = \int_{0}^{T}a(\omega, s)\cdot dW_s + \frac{1}{2}\int_{0}^{T}\|a(\omega, s)\|^2 ds~.
\end{eqnarray}
  \end{lemma}
\noindent With the above two key lemmas, we are able to derive our main theorem. 
\begin{theorem}[Continuous ELBO of DiffFlow]
Let $\{\hat{x}(t)\}_{t\in [0, T]}$ be a stochastic processes defined by (\ref{NSDE}) with initial distribution $\hat{x}(0)\sim q_0(x) $.  The marginal distribution of $\hat{x}(t)$ is denoted by $q_t(x)$.  Then the log-likelihood of the terminal marginal distribution has the following lower bound,
\begin{eqnarray}
&&\log q_T(x)\geq   \mathbb{E}_{Y_T}\left[\log q_0(Y_T)\Bigg|  Y_0 = x\right] +\frac{1}{2}\int_{0}^{T}\sigma^2(T-s)\mathbb{E}_{Y_s|Y_0=x}\left[\left\| \nabla\log p(Y_s|Y_0=x)\right\|_2^2 \right]ds \nonumber\\
&& -  \frac{1}{2}\int_{0}^{T}\mathbb{E}_{Y_s|Y_0=x}\left[\left\| \frac{c(Y_s, T-s;\theta_s)}{\sigma(T-s)}-\sigma(T-s)\nabla\log p(Y_s|Y_0=x)\right\|_2^2 \right]ds~.
\end{eqnarray}
where 
\begin{eqnarray}
&& dY_s = \sigma(T-s)d\widetilde{W}_s ~,\nonumber
\end{eqnarray}
and
$$c(x, t;\theta_t) = f(x, t) + \frac{g^2(t)}{2}s_{\theta_c^t}(x, t) + \left(\beta(t)-\frac{g^2(t)}{2}\right)\nabla d_{\theta_D^t}(x, t) ~,$$ and $$\sigma^2(t)= g^2(t)-\lambda^2(t)~.$$

\begin{proof}
By Fokker-Planck equation, we have
\begin{eqnarray}
&&\frac{\partial q_t(x)}{\partial t} +\nabla\cdot c(x, t;\theta_t)q_t(x) +c(x, t;\theta_t)\cdot\nabla q_t(x) +\frac{\sigma^2(t)}{2}\nabla\cdot\nabla q_t(x)=0~
\end{eqnarray}
where $$c(x, t;\theta_t) = f(x, t) + \frac{g^2(t)}{2}s_{\theta_c^t}(x, t) + \left(\beta(t)-\frac{g^2(t)}{2}\right)\nabla d_{\theta_D^t}(x, t) ~,$$ and $$\sigma^2(t)= g^2(t)-\lambda^2(t)~.$$
Let the time-reversal distribution $v_{t}(x) = q_{T-t}(x)$ for $0\leq t\leq T$, then $v_{t}(x)$ satisfies the following PDE,
\begin{eqnarray}
&&\frac{\partial v_t(x)}{\partial t} -\nabla\cdot c(x, T-t;\theta_s)v_t(x) -c(x, T-t;\theta_s)\cdot\nabla v_t(x) -\frac{\sigma^2(T-t)}{2}\nabla\cdot\nabla v_t(x)=0~.
\end{eqnarray}
By Feymann-Kac formula, we have
\begin{eqnarray}
&&q_T(x) = v_0(x) = \mathbb{E}\left[  q_0(Y_T)\exp\left(-\int_0^T \nabla\cdot c(Y_s, T-s;\theta_s) ds\right)   \Bigg| Y_0 = x\right]
\end{eqnarray}
where $Y_s$ is a diffusion process solving
\begin{eqnarray}
&& dY_s = -c(X_s, T-s;\theta_s)ds + \sigma(T-s)dW_s~.
\end{eqnarray}
By Jensen's Inequality,
\begin{eqnarray}
&&\log q_T(x) = \log \mathbb{E}_{\mathbb{Q}}\left[  \frac{d\mathbb{P} }{d\mathbb{Q} } q_0(Y_T)\exp\left(-\int_0^T \nabla\cdot c(Y_s, T-s;\theta_s) ds\right)   \Bigg| Y_0 = x\right]\nonumber\\
&&\geq \mathbb{E}_{\mathbb{Q}}\left[  \log \frac{d\mathbb{P} }{d\mathbb{Q} } +\log q_0(Y_T) -\int_0^T \nabla\cdot c(Y_s, T-s;\theta_s) ds  \Bigg| Y_0 = x\right]~.
\end{eqnarray}
Now, if we choose 
\begin{eqnarray}
d\widetilde{W}_s = a(\omega, s)ds + dW_s
\end{eqnarray}
and $\mathbb{Q}$ as 
\begin{eqnarray}
&& \log\frac{d\mathbb{P} }{d\mathbb{Q} }(\omega) \nonumber\\
&& = \int_{0}^{T}a(\omega, s)\cdot dW_s + \frac{1}{2}\int_{0}^{T}\|a(\omega, s)\|^2 ds  \nonumber\\
&& =  \int_{0}^{T}a(\omega, s)\cdot (d\widetilde{W}_s -a(\omega, s)ds ) + \frac{1}{2}\int_{0}^{T}\|a(\omega, s)\|^2 ds  \nonumber\\
&& =    \int_{0}^{T}a(\omega, s)\cdot d\widetilde{W}_s - \frac{1}{2}\int_{0}^{T}\|a(\omega, s)\|^2 ds
\end{eqnarray}
Then $d\widetilde{W}_s$ is Brownian motion under $\mathbb{Q}$ measure and 
\begin{eqnarray}
&&\log q_T(x) \nonumber\\
&&\geq \mathbb{E}_{\mathbb{Q}}\left[ \int_{0}^{T}a(\omega, s)\cdot d\widetilde{W}_s - \frac{1}{2}\int_{0}^{T}\|a(\omega, s)\|^2 ds +\log q_0(Y_T) -\int_0^T \nabla\cdot c(Y_s, T-s;\theta_s) ds  \Bigg| Y_0 = x\right] \nonumber\\
&& = \mathbb{E}_{\mathbb{Q}}\left[ -\frac{1}{2}\int_{0}^{T}\|a(\omega, s)\|^2 ds +\log q_0(Y_T) -\int_0^T \nabla\cdot c(Y_s, T-s;\theta_s) ds  \Bigg| Y_0 = x\right] \nonumber\\
&& = \mathbb{E}_{Y_T}\left[\log q_0(Y_T)\Bigg|  Y_0 = x\right] - \mathbb{E}_{\mathbb{Q}}\left[ \frac{1}{2}\int_{0}^{T}\left[\|a(\omega, s)\|^2  +  \nabla\cdot c(Y_s, T-s;\theta_s) \right]ds  \Bigg| Y_0 = x\right] ~.
\end{eqnarray}
Furthermore, we have
\begin{eqnarray}
&& dY_s = -c(Y_s, T-s;\theta_s) ds + \sigma(T-s)dW_s \nonumber\\
&& = -(c(Y_s, T-s;\theta_s) + \sigma(T-s)a(\omega, s))ds + \sigma(T-s)d\widetilde{W}_s  
\end{eqnarray}
By choosing appropriate $a(\omega, s)$, we can obtain a trainable ELBO. In particular, we choose
\begin{eqnarray}
&& a(\omega, s) = - c(Y_s, T-s;\theta_s) / \sigma(T-s)~.
\end{eqnarray}
Then we have
\begin{eqnarray}
&& dY_s = \sigma(T-s)d\widetilde{W}_s  ~.
\end{eqnarray}
and
\begin{eqnarray}
&&\log q_T(x) \nonumber\\
&&\geq  \mathbb{E}_{Y_T}\left[\log q_0(Y_T)\Bigg|  Y_0 = x\right] -  \frac{1}{2}\int_{0}^{T}\mathbb{E}_{Y_s}\left[\left(\frac{\|c(Y_s, T-s;\theta_s)\|^2}{\sigma^2(T-s)}  +  \nabla\cdot c(Y_s, T-s;\theta_s) \right)  \Bigg| Y_0 = x\right]ds \nonumber\\
&& = \mathbb{E}_{Y_T}\left[\log q_0(Y_T)\Bigg|  Y_0 = x\right] +\frac{1}{2}\int_{0}^{T}\sigma^2(T-s)\mathbb{E}_{Y_s|Y_0=x}\left[\left\| \nabla\log p(Y_s|Y_0=x)\right\|_2^2 \right]ds \nonumber\\
&& -  \frac{1}{2}\int_{0}^{T}\mathbb{E}_{Y_s|Y_0=x}\left[\left\| \frac{c(Y_s, T-s;\theta_s)}{\sigma(T-s)}-\sigma(T-s)\nabla\log p(Y_s|Y_0=x)\right\|_2^2 \right]ds~.
\end{eqnarray}
Then we can obtain the objective of the maximum likelihood inference by jointly training a weighted composite network $c(x, t;\theta_t) = f(x, t) + \frac{g^2(t)}{2}s_{\theta_c^t}(x, t) + \left(\beta(t)-\frac{g^2(t)}{2}\right)\nabla d_{\theta_D^t}(x, t)$ to be some weighted version of denoising score matching. 
\end{proof}
\end{theorem}

\section{Related Work}
Our work is inspired from recent series of work on relating the training dynamics of GANs and SDMs to the particle evolution of ordinary or stochastic differential equations, see \cite{gao2019deep, song2019generative, songscore, song2020denoising, karras2022elucidating, yi2023monoflow} and references therein.
\noindent The training dynamics of vanilla GANs \cite{goodfellow2020generative} is highly unstable due to the minimax objective. Since solving an optimal discriminator yields the some divergence minimization objective for GANs, and the divergence minimization naturally yields the particle algorithms and gradient flows, several subsequent works try to improve the stability of GAN training from the perspective of particle gradient flow, which avoids the minimax training framework \cite{nowozin2016f, arjovsky2017wasserstein, gulrajani2017improved, gao2019deep, yi2023monoflow}. The key idea is that the evolution of particles can be driven by the gradient flow of some distance measure between probability distributions, such as KL divergence or Wasserstein distance. Hence, the evolution dynamics can be an ODE and the driven term of ODE is determined by the functional gradient of the distance measure: for KL divergence, the functional gradient is the gradient field of the logistic classifier \cite{gao2019deep, yi2023monoflow}, which plays the role of the discriminator from the perspective of vanilla GANs. 

\noindent The earliest development of diffusion models are mainly focused on learning a series of markov transition operators that maximize the ELBO \cite{sohl2015deep, ho2020denoising}. In a parallel development, \cite{song2019generative, song2020improved} propose a series of score-based generative models based on multiple levels of denoising score matching. Until $2020$, the seminar work by \cite{songscore} figures out that diffusion models are essentially score-based generative models with score matching on multiple noise levels of corrupted target distributions. \cite{songscore} further shows that the sampling dynamics of diffusion models can be modelled as stochastic differential equations with scores of the noise-corrupted target as drift term.  Since then, we can name diffusion models as score-based diffusion models. 

\noindent Despite the training dynamics of both GANs and score-based diffusion models can be modelled by particle algorithms whose dynamics is described by a respective differential equation, there lacks a unified differential equation that can describe the dynamics of both. Our main contribution is to propose such an SDE that enables us build a continuous spectrum that unifies GANs and diffusion models.

\section{Conclusion}
We design a unified SDE ``DiffFlow'' that unifies the particle dynamics of the Langevin algorithm, GANs, diffusion SDEs, and diffusion ODEs. Our framework provides a continuous spectrum beyond SDMs and GANs, yielding new generative algorithms such as diffusion-GANs and SLCD.  We provide the asymptotic convergence analysis of the DiffFlow and show that we can perform the maximal likelihood inference for both GANs and SDMs under the current SDEs framework. However, the objective for maximal likelihood inference requires the joint training the score network and discriminator network to be some weighted version of denoising score matching, which would be hard to implement efficiently. It would be interesting to further explore how to choose a better reference measure in the Girsanov change of measure theorem to achieve a simpler trainable ELBO for maximal likelihood inference in DiffFlow.

\bibliographystyle{apalike}
\bibliography{main}

\end{document}